\documentclass[10pt,twocolumn,letterpaper]{article}

\usepackage{temp}
\usepackage{times}
\usepackage{epsfig}
\usepackage{graphicx}
\usepackage{amsmath}
\usepackage{amssymb}
\usepackage{multirow}
\usepackage{array}

\usepackage[pagebackref=true,breaklinks=true,letterpaper=true,colorlinks,bookmarks=false]{hyperref}

\cvprfinalcopy 

\ifcvprfinal\fi
\begin{document}

\title{Recurrent Calibration Network for Irregular Text Recognition}

\author{Yunze Gao$^{1}$ \qquad Yingying Chen$^{1}$ \qquad Jinqiao Wang$^{1}$ \\
Zhen Lei$^{1}$ \qquad Xiao-Yu Zhang$^{2}$ \qquad Hanqing Lu$^{1}$\\
$^{1}$National Laboratory of Pattern Recognition, Institute of Automation, \\
Chinese Academy of Sciences, Beijing, China\\
$^{2}$Institute of Information Engineering, Chinese Academy of Sciences, Beijing, China \\
{\tt\small \{yunze.gao, yingying.chen, jqwang, zlei, luhq\}@nlpr.ia.ac.cn }\\
{\tt\small  zhangxiaoyu@iie.ac.cn}}

\maketitle

\begin{abstract}
Scene text recognition has received increased attention in the research community. Text in the wild often possesses irregular arrangements, typically including perspective text, curved text, oriented text. Most existing methods are hard to work well for irregular text, especially for severely distorted text. In this paper, we propose a novel Recurrent Calibration Network (RCN) for irregular scene text recognition. The RCN progressively calibrates the irregular text to boost the recognition performance. By decomposing the calibration process into multiple steps, the irregular text can be calibrated to normal one step by step.
Besides, in order to avoid the accumulation of lost information caused by inaccurate transformation, we further design a fiducial-point refinement structure to keep the integrity of text during the recurrent process. Instead of the calibrated images, the coordinates of fiducial points are tracked and refined, which implicitly models the transformation information. Based on the refined fiducial points, we estimate the transformation parameters and sample from the original image at each step. In this way, the original character information is preserved until the final transformation. Such designs lead to optimal calibration results to boost the performance of succeeding recognition. Extensive experiments on challenging datasets demonstrate the superiority of our method, especially on irregular benchmarks.
\end{abstract}

\section{Introduction}

Scene text recognition has drawn remarkable attention in computer vision due to its importance in various real world applications, such as scene understanding, card information entry, street sign reading and so on. Benefiting from recent advancements of deep learning, reading text in natural images has experienced a rapid evolution during the past few years. Despite significant advances, scene text recognition in unconstrained conditions still remains as a challenging problem due to the complex situations such as blurring, distortion, orientation and uneven lighting.

\begin{figure}
\centering
\includegraphics[width=7.5cm]{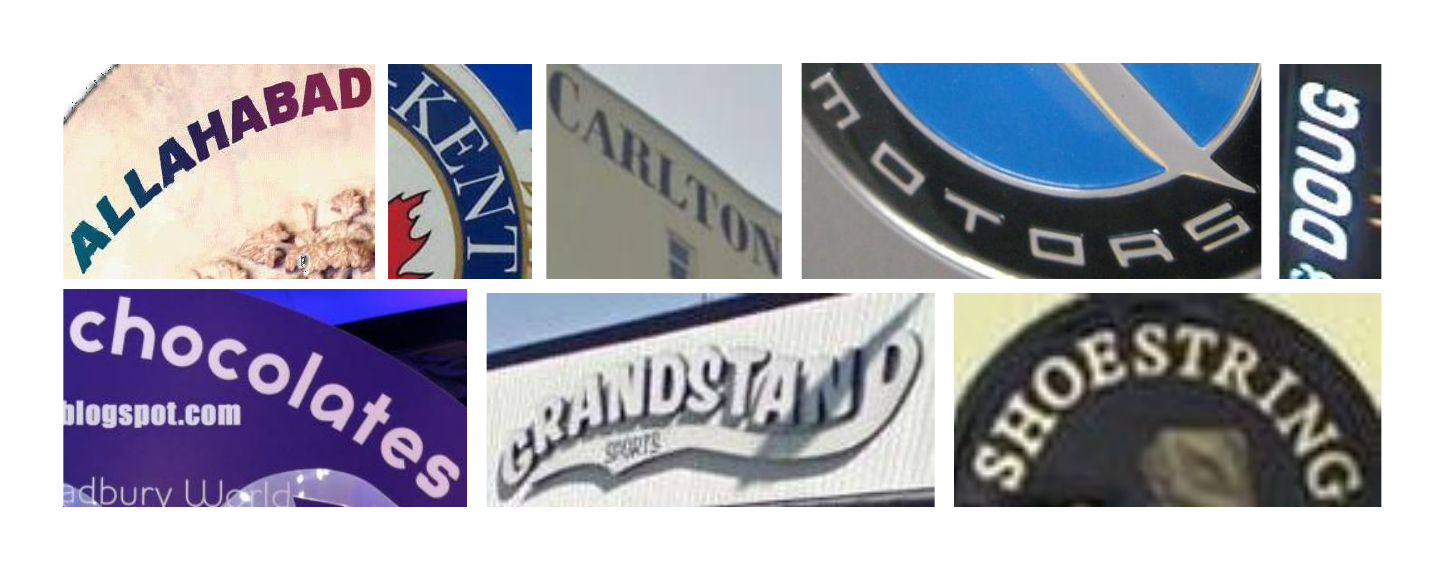}
\caption{Examples of irregular text in natural images.}
\label{fig:picture004}
\end{figure}

\begin{figure*}
\centering
\includegraphics[width=17.5cm]{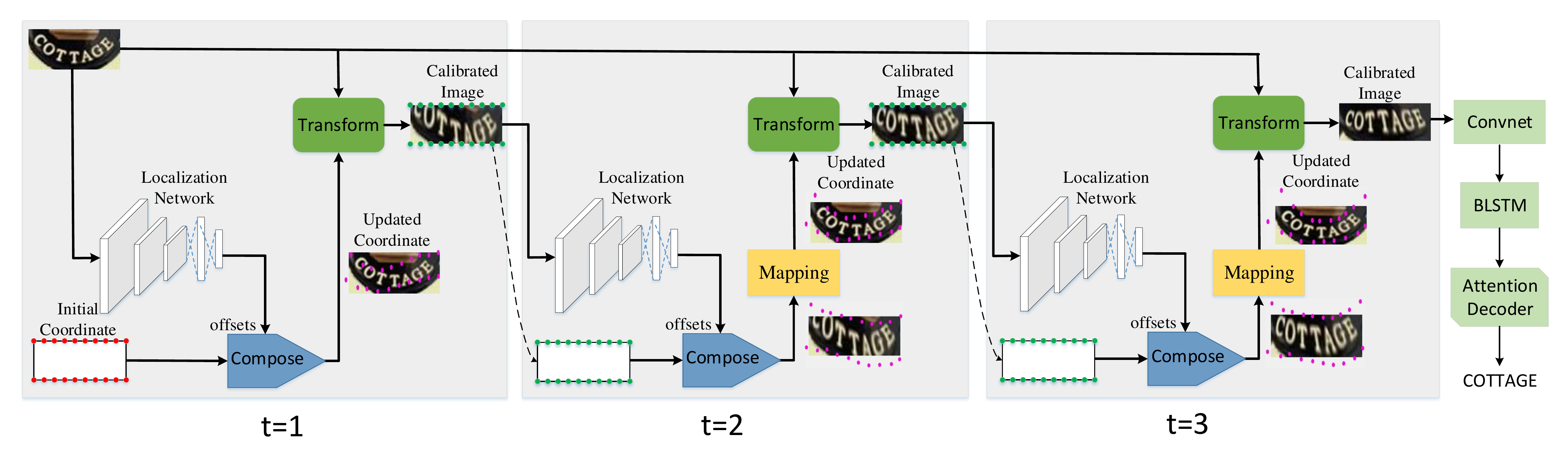}
\caption{Overview of the Recurrent Calibration Network (RCN) for irregular text recognition. Our RCN recurrently refines the coordinates of fiducial points and transforms the original input image. The dotted lines represent the coordinates delivery.}
\label{fig:picture001}
\end{figure*}

Irregular text frequently appears in natural scenes, owing to curved character placement, perspective distortion, etc., as shown in Figure \ref{fig:picture004}. Recognizing text with arbitrary shapes is an extremely difficult task because of unpredictable changeful text layouts. Most existing approaches mainly focus on regular text recognition, which are difficult to be generalized to distorted text. Recently, some attempts have been made towards irregular text recognition. Yang \emph{et al.} \cite{yang2017learning} utilized a 2D attention mechanism to focus on each character and introduced an auxiliary dense character detection task to encourage the learning of text-specific visual representations. However, this method used an exhausting multi-task learning strategy and the inaccurate attention regions will cause recognition errors. Cheng \emph{et al.} \cite{cheng2017arbitrarily} proposed the arbitrary orientation network (AON) to extract scene text features in four directions and adopted a weighting mechanism to combine the four feature sequences of different directions. In order to extract features with the same dimension in four directions, AON has to resize the text image into a square shape. However, scene text generally has various aspect ratios, the strategy of scaling to a square will severely destroy the aspect ratio of the text line, especially for long text. Shi \emph{et al.} \cite{shi2018aster} applied a spatial transformation prior to recognition to transform the input image and rectify the text in it. The Spatial Transformer Network (STN) framework with Thin-Plate Spline (TPS) transformation is utilized to perform text rectification. Although \cite{shi2018aster} has shown impressive results on irregular benchmarks, we observe that the rectified images may still have distortions or lose some character information, especially for severely distorted text, which results in mistaken recognition results.

In this paper, we design a novel Recurrent Calibration Network (RCN) to progressively calibrate the irregular text to boost the recognition performance. The recurrent structure iteratively refines the geometric transformation of irregular text under the same parametric capacity. In each iteration, the residual between the previous and current geometric transformation fields is estimated based on the previously calibrated image to get one step closer to the optimal one. In this way, the difficulty of each step is intrinsically relieved and the severe distortions can be eliminated in a progressive manner. Therefore, such design is capable of effectively improving the robustness of the model to large variations of text. Besides, we observe that spatial transformation on output image of the previous step cannot restore the missing character information, and the incomplete appearances will cause recognition errors as well. Therefore, we elaborate a fiducial-point refinement structure to keep the integrity of text during the recurrent process. Instead of the calibrated images, the coordinates of fiducial points are tracked and transmitted during multiple iterations. At each step, the localization network predicts the coordinate offsets with respect to the previous positions, which implicitly reflects the residual spatial transformation. Furthermore, we map the coordinates of fiducial points back to the original input image and sample from the original. In this way, while the coordinates fall outside the calibrated image, mapping back to the original means compensating some missing information. Our method can effectively calibrate the irregular text while preserving the original character information in multiple calibrations. The calibration network is jointly optimized with the recognition network under the same objective in an end-to-end scheme. Therefore, our RCN can automatically learn the optimal transformation for the following recognition task.

The main contributions are summarized as follows:

(1) We propose a Recurrent Calibration Network (RCN) to progressively calibrate the irregular text to boost the recognition performance.

(2) We design a fiducial-point refinement structure to keep the integrity of text during the recurrent process, which avoids the accumulation of missing information in the scenario of iterative calibrations.

(3) Our RCN achieves superior performance compared with the state-of-the-art methods on the challenging datasets, especially on irregular benchmarks.

\section{Related Work}

Scene text recognition has been widely researched and numerous methods are proposed in recent years. Traditional methods recognized scene text in a character-level manner, which first performed detection to generate multiple candidates of character locations, then applied a character classifier for recognition. Wang \emph{et al.} \cite{wang2011end} detected each character by sliding window, and recognized it with a character classifier trained on the HOG descriptors. Bissacco \emph{et al.} \cite{bissacco2013photoocr} designed a fully connected network to extract character feature representations, then used a language model to recognize characters. However, the performance of these methods is limited due to the inaccurate character detector. To be free from this problem, some methods directly learned the mapping between the entire word images and target strings. For example, Jaderberg \emph{et al.} \cite{jaderberg2016reading} assigned a class label to each word in a pre-defined lexicon and performed a 90k-class classification with CNN. Rodriguez-Serrano \emph{et al.} \cite{rodriguez2015label} formulated the scene text recognition as a retrieval problem, which embedded word labels and word images into a common Euclidean space and found the closest word label in this space.

With the successful application of recurrent neural network (RNN) in sequence recognition, some researchers \cite{shi2017end,he2016reading,shi2016robust,lee2016recursive} developed sequence-based methods and combined convolutional neural network (CNN) and RNN to encode the feature representations of word images. Shi \emph{et al.} \cite{shi2017end} and He \emph{et al.} \cite{he2016reading} both used the Connectionist Temporal Classification (CTC) \cite{graves2006connectionist} loss to calculate the conditional probabilities between the outputs of RNN and the target sequences. After that, Shi \emph{et al.} \cite{shi2016robust} and Li \emph{et al.} \cite{lee2016recursive} introduced an attention mechanism to adaptively weight the features and select the most relevant feature representations in RNN-based decoder. In order to eliminate attention drift problem, Cheng \emph{et al.} \cite{cheng2017focusing} employed a focusing attention mechanism to automatically adjust the attention weights. Bai \emph{et al.} \cite{bai2018edit} proposed the edit probability to estimate the probability of generating a string while considering possible occurrences of missing or superfluous characters. Although these approaches have shown promising results, they cannot effectively handle with the irregular text. The main reason is that word images are encoded into 1D feature sequences, but the irregular text is not horizontally arranged.

The researches on irregular text recognition are relatively sparse. Yang \emph{et al.} \cite{yang2017learning} adopted a 2D soft attention mechanism to focus on individual character at each step and introduced an auxiliary character detection task to learn text-specific features. Although this method recognized text in a 2D space, it used the multi-task learning framework and needed character-level annotations. Liu \emph{et al.} \cite{liu2018char} presented to remove the distortion of scene text by detecting and rectifying individual character. However, the detection errors will affect the performance of subsequent rectification and recognition. Cheng \emph{et al.} \cite{cheng2017arbitrarily} proposed that the visual representation of irregular text can be described as the combination of features in four directions. This approach is able to effectively capture the deep features of irregular text, but the strategy of scaling word images to square will severely destroy the aspect ratio of text lines, especially for long text. Shi \emph{et al.} \cite{shi2018aster} introduced a spatial transformation network with thin-plate spline (TPS) transformation to calibrate irregular text into regular one, then recognized the calibrated text with attention-based framework. Although considerably improving the performance for irregular text recognition, it is still difficult to precisely locate the fiducial points which tightly bound the text region, especially for severely distorted text. This leads to errors in parameters estimation of the TPS transformation, and hence the deformation of scene text.

Different from the existing methods, we design a novel Recurrent Calibration Network (RCN) to calibrate irregular text in a progressive manner. The calibration process is decomposed into multiple steps and the calibration results are iteratively refined. Different steps work together to eliminate text distortion for better recognition, thus the difficulty of each step is greatly relieved and large distortion also can be effectively removed.

\section{The proposed Approach}

The overview of our Recurrent Calibration Network (RCN) for irregular text recognition is shown in Figure \ref{fig:picture001}. The irregular text is progressively calibrated to regular one, which serves as the input of subsequent recognition network. During calibration process, the distortions are eliminated step by step and the recurrent framework maintains the same parametric capacity. Besides, the fiducial-point refinement structure transmits and refines the coordinates of fiducial points during the recurrent process. At each step, the network first predicts the coordinate offsets, then gets the updated coordinates and projects them into the original input image. After that, we can estimate the TPS parameters and sample from the original image, which effectively keeps the integrity of text. RCN is able to deal with distorted text with various orientations and shapes, including severely distorted text.

\subsection{Recurrent Calibration}

The spatial transformer \cite{jaderberg2015spatial} is a learnable module which explicitly allows spatial manipulations. This can be mathematically written as
\begin{equation}
I_{out}=I_{in}(T), \text{where } T=f(I_{in})
\end{equation}
The function $f$ is parameterized as a learnable localization network, which predicts the transform parameters from the input image. Usually, it is common to calibrate scene text with spatial transform network. 

However, the single-step calibration often fails to fully remove geometrical distortions, and could lead to text content loss, making unfavorable effects on the following recognition. Therefore, we design a recurrent structure to decompose the calibration process into multiple progressive steps, which mitigates the difficulty of each step. Different steps work together to calibrate irregular text for better recognition. In each iteration, the calibrated result is further refined by feeding it into the pipeline again, forming a recurrent structure. Therefore, we can refine the calibration results iteratively under the same parametric capacity. Moreover, we note that the grid generator and the sampler in STN can be combined to be a single transform function. Denote the transformation parameters estimation as $E$, and the spatial transformation operation as $R$, we have the following structure:
\begin{equation}
T_{t} = E(I_{t})
\end{equation}
\begin{equation}
I_{t+1} = R(I_{t}, T_{t})
\end{equation}
where $t$ represent the $t$-$th$ iteration, and $I_{0}$ is the original input image. In this way, large variations can be progressively handled to match with the succeeding recognition task.

For each transformation, we predict a set of fiducial points and calculate the TPS transformation parameters based on them. Specifically, the base fiducial points $C$ on target image $I_{out}$ are defined to evenly distribute along the top and bottom image borders, which are always constant. In the feed-forward process, the localization network regresses the coordinates of fiducial points $C'$ on input image $I_{in}$. Assuming K fiducial points on both $I_{in}$ and $I_{out}$, the parameters of the TPS transformation is represented by a $2\times (K+3)$ matrix:
\begin{equation}
T=[C'  \quad  0^{2\times 3}]\Delta_{C}^{-1}
\label{eq:eq001}
\end{equation}
where $\Delta_{C}$ is a $(K+3)\times (K+3)$ matrix calculated from C:
\begin{equation}
\Delta_{C}=
    \begin{bmatrix}
    1^{1\times K} & 0 & 0 \\
    C & 0 & 0 \\
    \hat{C} & 1^{K\times 1} & C^\top
    \end{bmatrix}
\end{equation}
where $\hat{C}$ is a $K \times K$ matrix comprising $\hat{c}_{a,b}=\phi(\| c_{a}-c_{b} \|)$ and $\phi(r)=r^{2}\log(r)$. Given a point $p = [x_{p}  \quad y_{p}]^\top$ on $I_{out}$, TPS finds its corresponding point $p'$ by linearly projection:
\begin{equation}
    p'=T
    \begin{bmatrix}
    1 \\
    p \\
    \phi (\| p-c_{1} \|)\\
    \phi (\| p-c_{K} \|)
    \end{bmatrix}
\label{eq:eq002}
\end{equation}
In the transform module, a grid $\mathcal{P}$ on $I_{in}$ is obtained by iterating over all points on $I_{out}$, and the calibrated image is generated by bilinear interpolation based on $\mathcal{P}$.

\subsection{Fiducial-point Refinement}

\begin{figure}
\centering
\includegraphics[width=7cm]{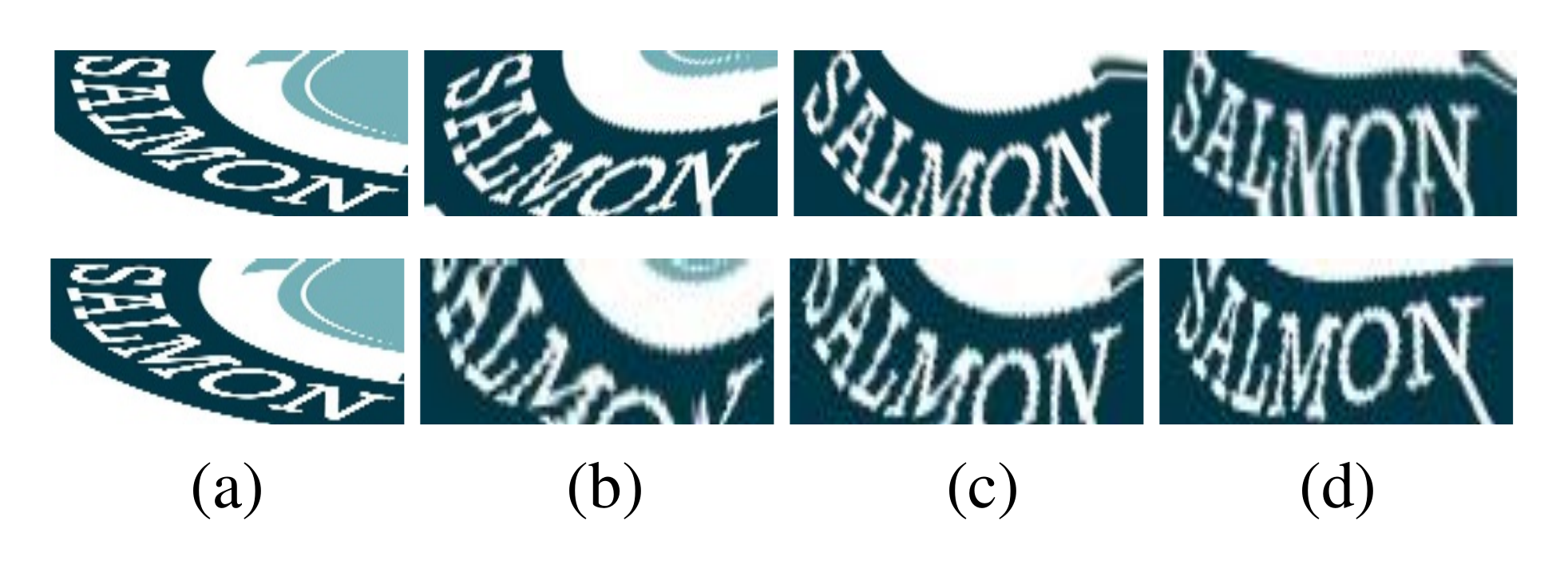}
\caption{The comparison of two recurrent methods, the top and bottom are the direct iterations structure and fiducial-point refinement structure, respectively. (a) The original input images. (b) The calibrated images in the first iteration. (c) The calibrated images in the second iteration. (d) The calibrated images in the last iteration. Direct iterations structure discards the information outside the calibrated images, while the missing information can be recovered in the fiducial-point refinement structure.}
\label{fig:picture002}
\end{figure}

As observed in Figure \ref{fig:picture002}, the iterative refinements can gradually calibrate the irregular text to be more beneficial to recognition, however, the missing information caused by inaccurate transformation cannot be restored in the direct iterations structure. This leads to incomplete character appearances and thus results in the recognition errors. We analyze that the output image samples only from the calibrated image and pixel information outside the region is discarded. In the top row of Figure \ref{fig:picture002}, such effect is visible while requiring pixel information outside the calibrated images. In the scenario of iterative calibrations, the effects of missing information are accumulated during multiple transformations.

To remedy this issue, we design a fiducial-point refinement structure in the network to keep the integrity of text in the recurrent process. It is worth noting that the transformation parameters are only defined by the fiducial points. Therefore, we advocate that the transformation information is transmitted through the fiducial points, rather than being discarded after transformation. During multiple iterations, the coordinates of fiducial points are kept track of and refined. Based on the refined fiducial points, we estimate the transformation parameters and sample from the original image at each step. Therefore, the original character information is preserved until the final transformation. In order to ease the training of networks, the localization network just predicts the coordinates offsets. It is easier to optimize the residual coordinates than to optimize the original coordinates. Furthermore, the offsets and the previous coordinates are composed to describe the current positions of fiducial points. However, based on the calibrated images, the generated offsets and coordinates are both on the calibrated images. In order to sample from the original image, we need to map the coordinates back to the original image. Especially, at the first step, the input is the original image, so the mapping is an identity transformation and can be omitted. It also should be noted that the fiducial points on original image are mapped again to the top and bottom image borders on calibrated image after each transformation. Therefore, the previous fiducial points are always the canonical forms (see green points in Fig \ref{fig:picture001}). In $t$-$th$ recursion, assuming that the offsets of fiducial points are denoted as $O_{t}$, and the coordinates of fiducial points on original image and calibrated image are $F^{ori}_{t}$ and $F^{cal}_{t}$ separately. The current coordinates of fiducial points on calibrated images are generated as the sum of the offsets and the previous coordinates:
\begin{equation}
F^{cal}_{t} = F^{cal}_{t-1}+O_{t}
\end{equation}
where
\begin{equation}
O_{t} = L(I_{t-1})
\end{equation}
$I_{t-1}$ is the output calibrated image of the previous step, and $L$ is the localization network. Then the current coordinates are mapped back to the original image, which serve as the updated coordinates. The mapping operation $M$ is the same as the definition in Eq. \ref{eq:eq002}, and the formulation is shown as
\begin{equation}
F^{ori}_{t} = M(F^{cal}_{t}, T_{t-1})
\end{equation}

Based on the coordinates of the updated fiducial points $F^{ori}_{t}$, the transformation parameters $T_{t}$ can be estimated as in Eq. \ref{eq:eq001} and the next calibrated image can be sampled from the original image $I_{ori}$:
\begin{equation}
I_{t} = R(I_{ori}, T_{t})
\end{equation}

In this way, the integrity of text is effectively kept because pixel information outside the calibrated image is also preserved until the final transformation. Therefore, our network can progressively calibrate the irregular text without any character information loss. Besides, all modules of the calibration process are differentiable, allowing for backpropagation within an end-to-end learning framework. Moreover, the calibration process focuses on the text region, which implicitly models the attention mechanism. Localizing the text region accurately not only can achieve satisfactory calibration, but also effectively removes the background noises.

\subsection{Recognition Network}

For the recognition network, we use the attention-based encoder-decoder pipeline as in \cite{shi2018aster}. First, the encoder extracts a sequence of feature vectors $H=(h_{1},\dots,h_{n})$ through CNN-LSTM structure. Then attention decoder recurrently generates the character sequence $y=(y_{1},\dots,y_{m})$. At step $i$, the decoder dynamically weights the image feature and selects the most relevant contents to generate the probability distribution $p(y_{i})$. Given the last RNN hidden state $s_{i-1}$ and feature sequence $H$, the attention weights can be obtained by scoring each element in feature sequence separately:
\begin{equation}
e_{i,j} = v^{T}tanh(Ws_{i-1}+Uh_{j}+b)
\end{equation}
\begin{equation}
\alpha_{i,j} = \frac{\exp(e_{i,j})}{\sum_{j=1}^{n}\exp(e_{i,j})}
\end{equation}
Then we can obtain the weighted sum of sequential feature vectors, which focuses on the most relevant features:
\begin{equation}
g_{i} = \sum_{j=1}^{n}\alpha_{i,j}h_{j}
\end{equation}
After that, the RNN hidden state is updated and the probability distribution $p(y_{i})$ is estimated as follows:
\begin{equation}
s_{i} = RNN(y_{i-1}, s_{i-1}, g_{i})
\end{equation}
\begin{equation}
p(y_{i}) = softmax(V^{T}s_{i})
\end{equation}
Above, $W$, $U$, $V$, $v$, $b$ are the learnable parameters. Following \cite{shi2018aster}, we exploited a bidirectional decoder, which consists of two decoders in opposite directions.

\subsection{Model Training}
Given the input image $I$ and corresponding ground truth $\hat{y} = (\hat{y_{1}}, \hat{y_{2}}, \cdots, \hat{y_{m}})$, the objective function is formulated by considering both the left-to-right decoder and the right-to-left decoder as follows:
\begin{equation}
L(\theta)=-\frac{1}{2} \sum_{i=1}^{m}\ (log p_{l2r}(\hat{y_{i}}|I)+\log p_{r2l}(\hat{y_{i}}|I))
\end{equation}
in which, $\theta$ denotes the parameters of both calibration network and recognition network, $p_{l2r}$ and $p_{r2l}$ are the output probability distributions of decoders in left-to-right and right-to-left order, respectively. Through the recurrent structure, we can perform multiple calibrations under the same parametric capacity. Furthermore, the calibration network and recognition network are optimized together under the same recognition loss, so the calibration network is encouraged to transform the irregular text to best match the succeeding recognition network.

\section{Experiments}

\begin{table*}
  \centering
  \caption{Scene text recognition accuracies on irregular datasets. ``50'' and ``Full'' represent the size of lexicon used for lexicon-based recognition, and ``None'' represents lexicon-free recognition.}
  \begin{tabular}
  {c|c|c|c|c|c|p{2.4cm}<{\centering}|p{2.4cm}<{\centering}}
  \hline
  \multirow{2} * {Methods} & \multicolumn{3}{c|} {\multirow{2}*{SVT-Perspective}} & \multirow{2}*{CUTE80} & \multirow{2}*{IC15} & {Total-Text  (multi-oriented)} & {Total-Text  (curved)} \\
  \cline{2-8}  & 50 & Full & None & None & None & None & None   \\
  \hline\
  ABBYY\cite{wang2011end} & 40.5 & 26.1 & - & - & - & - & -  \\
  Mishra \emph{et al}.\cite{mishra2012top} & 45.7 & 24.7 & - & - & - & - & - \\
  Phan \emph{et al}. \cite{quy2013recognizing} & 75.6 & 67.0 & - & - & - & - & - \\
  Shi \emph{et al}.\cite{shi2017end} & 92.6 & 72.6 & 66.8 & 54.9 & - & - & - \\
  Yang \emph{et al}.\cite{yang2017learning} & 93.0 & 80.2 & 75.8 & 69.3 & - & - & -\\
  Shi \emph{et al}.\cite{shi2016robust} & 91.2 & 77.4 & 71.8 & 59.2 & - & - & - \\
  Liu \emph{et al}. \cite{liu2018char} & - & - & 73.5 & - & - & - & -  \\
  Cheng \emph{et al}. \cite{cheng2017focusing} & 92.6 & 81.6 & 71.5 & 63.9 & 66.2 & - & -  \\
  Cheng \emph{et al}. \cite{cheng2017arbitrarily} & 94.0 & 83.7 & 73.0 & 76.8 & 68.2 & - & -  \\
  Shi \emph{et al}. \cite{shi2018aster} & - & - & 78.5 & 79.5 & 76.1 & - & - \\
  \hline
  baseline & 93.4 & 89.7 & 77.0 & 82.3 & 72.0 & 72.3 &  56.4\\
  RCN(ours) & \textbf{95.0} & \textbf{91.2} & \textbf{80.6} & \textbf{88.5} & \textbf{77.1} & \textbf{76.3} & \textbf{66.7} \\
  \hline
\end{tabular}
\label{tab:001}
\end{table*}

In this section, we describe the details of experimental settings and evaluate the effectiveness of our method. We compare the performance of our RCN with the other approaches on both regular and irregular datasets.

\subsection{Datasets}

\begin{itemize}
\item \textbf{Street View Text Perspective (SVT-P)} \cite{quy2013recognizing} contains 639 cropped word images which are captured from the side-view angles in Google Street View. Most of them suffer from severely perspective distortion. Each image is specified with a 50 words lexicon and a full lexicon.
\item \textbf{CUTE80} \cite{risnumawan2014robust} contains 288 cropped word images for testing, which is specially collected for evaluating the performance of curved text recognition. No lexicon is provided.
\item \textbf{ICDAR 2015} \cite{karatzas2015icdar} contains 2077 word images including plenty of irregular text, which are taken from Google Glasses. For fair comparison, we discard the images that contain non-alphanumeric characters. No lexicon is specified.
\item \textbf{Street View Text} \cite{wang2011end} contains 647 word images which are collected from Google Street View. Each image is associated with a 50 words lexicon defined by \cite{wang2011end}. Many images suffer from low resolution, blur and noise.
\item \textbf{IIIT5K} \cite{mishra2012top} contains 3000 cropped word images collected from the Internet. Each image has a 50 words lexicon and a 1000 words lexicon.
\item \textbf{ICDAR 2003} \cite{lucas2005icdar} contains 860 cropped word images for testing. Following the evaluation protocol in \cite{wang2011end}, we recognize the images containing only alphanumeric characters with at least three characters. Each image is specified with a 50 words lexicon defined by \cite{wang2011end}. And a full lexicon consists of all the words that appear in the test set.
\item \textbf{ICDAR 2013} \cite{karatzas2013icdar} derives from the ICDAR 2003. Following \cite{shi2018aster}, we remove the images that contain non-alphanumeric characters, which results in 1015 cropped word images without any pre-defined lexicon.
\item \textbf{ToTal-Text} \cite{ch2017total} has annotated word images with three different text orientations including horizontal, multi-oriented, and curved text. We select the multi-oriented and curved text collections, which contain 480 and 971 images separately.
\end{itemize}

We use the synthetic dataset as the training data, including Synth90k released by Jaderberg \emph{et al.} \cite{jaderberg2014synthetic} and SynthText released by Gupta \emph{et al.} \cite{gupta2016synthetic} . Our model is evaluated on all other real-world test datasets without any finetuning.

\begin{table*}
  \centering
  \caption{Scene text recognition accuracies on regular datasets. ``50'', ``1000'' and ``Full'' represent the size of lexicon used for lexicon-based recognition, and ``None'' represents lexicon-free recognition.}

  \begin{tabular}{c|c|c|c|c|c|c|c|c|c}
  \hline
  \multirow{2} * {Methods} & \multicolumn{2}{c|} {SVT} & \multicolumn{3}{c|} {IIIT5k} & \multicolumn{3}{c|} {IC03} & IC13 \\
  \cline{2-10}  & 50 & None & 50 & 1k & None & 50 & Full & None & None \\
  \hline
  Wang \emph{et al}.\cite{wang2011end} & 57.0 & - & - & - & - & 76.0 & 62.0 & - & - \\
  Mishra \emph{et al}.\cite{mishra2012top} & 73.2 & - & 64.1 & 57.5 & - & 81.8 & 67.8 & - & - \\
  Wang \emph{et al}.\cite{wang2012end} & 70.0 & - & - & - & - & 90.0 & 84.0 & - & - \\
  Bissacco \emph{et al}.\cite{bissacco2013photoocr} & 90.4 & 78.0 & - & - & - & - & - & - & 87.6 \\
  Almaz\'an \emph{et al}.\cite{almazan2014word} & 89.2 & - & 91.2 & 82.1 & - & - & - & - & - \\
  Yao \emph{et al}.\cite{yao2014strokelets} & 75.9 & - & 80.2 & 69.3 & - & 88.5 & 80.3 & - & - \\
  Rodriguez-Serrano \emph{et al}.\cite{rodriguez2015label} & 70.0 & - & 76.1 & 57.4 & - & - & - & - & - \\
  Jaderberg \emph{et al}.\cite{jaderberg2014deep} & 86.1 & - & - & - & - & 96.2 & 91.5 & - & - \\
  Gordo \cite{gordo2015supervised} & 91.8 & - & 93.3 & 86.6 & - & - & - & - & - \\
  Jaderberg \emph{et al}.\cite{jaderberg2016reading} & 95.4 & 80.7 & 97.1 & 92.7 & - & 98.7 & \textbf{98.6} & 93.1 & 90.8 \\
  Jaderberg \emph{et al}.\cite{jaderberg2014deep2} & 93.2 & 71.7 & 95.5 & 89.6 & - & 97.8 & 97.0 & 89.6 & 81.8 \\
  Shi \emph{et al}.\cite{shi2017end} & 97.5 & 82.7 & 97.8 & 95.0 & 81.2 & 98.7 & 98.0 & 91.9 & 89.6 \\
  Lee \emph{et al}.\cite{lee2016recursive} & 96.3 & 80.7 & 96.8 & 94.4 & 78.4 & 97.9 & 97.0 & 88.7 & 90.0 \\
  He \emph{et al}.\cite{he2016reading} & 92.0 & - & 94.0 & 91.6 & - & 97.0 & 94.4 & - & - \\
  Wang and Hu\cite{wang2017gated} & 96.3 & 81.5 & 98.0 & 95.6 & 80.8 & 98.8 & 97.8 & 91.2 & - \\
  Cheng \emph{et al}. \cite{cheng2017focusing} & 97.1 & 85.9 & 99.3 & 97.5 & 87.4 & \textbf{99.2} & 97.3 & 94.2 & 93.3 \\
  Bai \emph{et al}. \cite{bai2018edit} & 96.6 & 87.5 & 99.5 & 97.9 & 88.3 & 98.7 & 97.9 & \textbf{94.6} & \textbf{94.4} \\
  Liu \emph{et al}. \cite{liu2018squeezedtext} & 96.1 & - & 96.9 & 94.3 & 86.6 & 98.4 & 97.9 & 93.1 & 92.7 \\
  Yang \emph{et al}.\cite{yang2017learning} & 95.2 & - & 97.8 & 96.1 & - & 97.7 & - & - & - \\
  Shi \emph{et al}.\cite{shi2016robust} & 95.5 & 81.9 & 96.2 & 93.8 & 81.9 & 98.3 & 96.2 & 90.1 & 88.6 \\
  Liu \emph{et al}. \cite{liu2018char} & - & 84.4 & - & - & 83.6 & - & - & 91.5 & 90.8  \\
  Cheng \emph{et al}. \cite{cheng2017arbitrarily} & 96.0 & 82.8 & 99.6 & 98.1 & 87.0 & 98.5 & 97.1 & 91.5 & - \\
  Shi \emph{et al}. \cite{shi2018aster} & 97.4 & \textbf{89.5} & 99.6 & 98.8 & 93.4 & 98.8 & 98.0 & 94.5 & 91.8 \\
  \hline
  baseline & 96.4 & 86.2 & 99.6 & 98.9 & 92.6 & 98.6 & 97.6 & 92.9 & 90.7 \\
  RCN(ours) & \textbf{97.7} & 88.6 & \textbf{99.6} & \textbf{98.9} & \textbf{94.0} & 99.0 & 97.9 & 93.6 & 93.2 \\
  \hline
\end{tabular}
\label{tab:002}
\end{table*}

\subsection{Implementation Details}

To validate the effectiveness of our method, we use the same network architecture and experimental settings with \cite{shi2018aster} to ensure fair comparison. We just replace the spatial transform process with our proposed recurrent calibration framework. When the number of iterations is one, the RCN degenerates to the base model \cite{shi2018aster}. Notice that the calibration network uses its own shared weight matrixes, so our RCN has the same parametric capacity as the base model. Moreover, we also explore the effect of the number of iterations. The input images are resized to 64$\times$256, and the 32$\times$64 downsampled images serve as the input of the localization network. After performing spatial transformation, the calibrated images have the size of 64$\times$256 in the middle steps and 32$\times$100 in the last step. Besides, we do not use any data augmentation. The model is trained with the ADADELTA \cite{zeiler2012adadelta} optimization method.

During the process of testing, we report the results for both lexicon-free and lexicon-based recognition. In lexicon-free setting, we directly select the most probable character at each decoding step. Then the bidirectional decoder generates two results and we choose the result with the higher probability. In the lexicon-based setting, we pick the nearest lexicon word with the generated string under the metric of edit distance.

Our network is implemented under the Pytorch \cite{ketkar2017introduction} framework. Most parts of our model are GPU-accelerated due to the CUDA backend. All the experiments are carried out on a workstation which has one Inter(R) Xeon(R) E5-2630 2.20Ghz CPU, an NVIDIA TITAN X GPU and 256GB RAM.

\subsection{Depth of Recurrence}

The recurrent structure progressively calibrates the irregular text and thus generates better calibrated images. We investigate the effect of the number of iterations and report the results in Table \ref{tab:003}. As the number of iterations increases, the recognition results also gradually perform better. In particular, the performance improvements on curved text benchmarks are remarkable, which suggests the significance of iterative calibrations in recognizing severely distorted text. Moreover, our network degenerates to \cite{shi2018aster} when the number of iterations is one. Compared with \cite{shi2018aster}, we do not reproduce their reported results and fall behind them. However, we still achieve better performance than \cite{shi2018aster}, which demonstrates the effectiveness of our designs. More than three iterations result in negligible effects and the number of iterations is set as three in the following experiments.

In addition, by comparing the RCN-3 and RCN-3 (w/o \emph{FP-R}), we can see that the fiducial-point refinement structure leads to significant performance promotion under the same number of iterations. The main reason is that the original character information is preserved and thus the missing information can be recovered. And the succeeding recognition network benefits from the integrity of text.

Furthermore, some examples are presented in Figure \ref{fig:picture003}. As observed, the text becomes more regular with the number of iterations increases. Besides, the lost character information in the previous step can be recovered in the subsequent processes. Therefore, the integrity of text is effectively preserved during the iterative calibrations. Our network not only transforms the text in the direction that is more beneficial to recognition, but also gradually removes the background noises.

\subsection{Performance on Irregular Benchmarks}

\begin{table*}
  \centering
  \caption{Lexicon-free results on several benchmarks with the different number of iterations. \emph{FP-R} represents the proposed fiducial-point refinement structure.}
  \begin{tabular}
  {c|c|c|c|c|c|c|c|p{2.3cm}<{\centering}|p{1.45cm}<{\centering}}
  \hline
  \multirow{2}*{Method} & \multirow{2}*{SVT} & \multirow{2}*{IIIT5k} & \multirow{2}*{IC03} & \multirow{2}*{IC13} & \multirow{2}*{SVT-P} & \multirow{2}*{CUTE80} & \multirow{2}*{IC15} & {Total-Text (multi-oriented)} & {Total-Text (curved)} \\
  \hline
  RCN-3(w/o \emph{FP-R}) & 87.5 & 92.4 & 93.5 & 91.5 & 78.1 & 86.1 & 74.1 & 73.8 & 65.2 \\
  RCN-1  & 86.2 & 92.6 & 92.9 & 90.7 & 77.0 & 82.3 & 72.0 & 72.3 & 56.4\\
  RCN-2 & 85.6 & 92.7 & 93.0 & 91.6 & 77.9 & 83.0 & 74.6 & 75.2 & 57.6 \\
  RCN-3 & \textbf{88.6} & \textbf{94.0} & \textbf{93.6} & \textbf{93.2} & \textbf{80.6} & \textbf{88.5} & \textbf{77.1} & \textbf{76.3} & \textbf{66.7}\\
  \hline
\end{tabular}
\label{tab:003}
\end{table*}

\begin{figure}
\centering
\includegraphics[width=8cm]{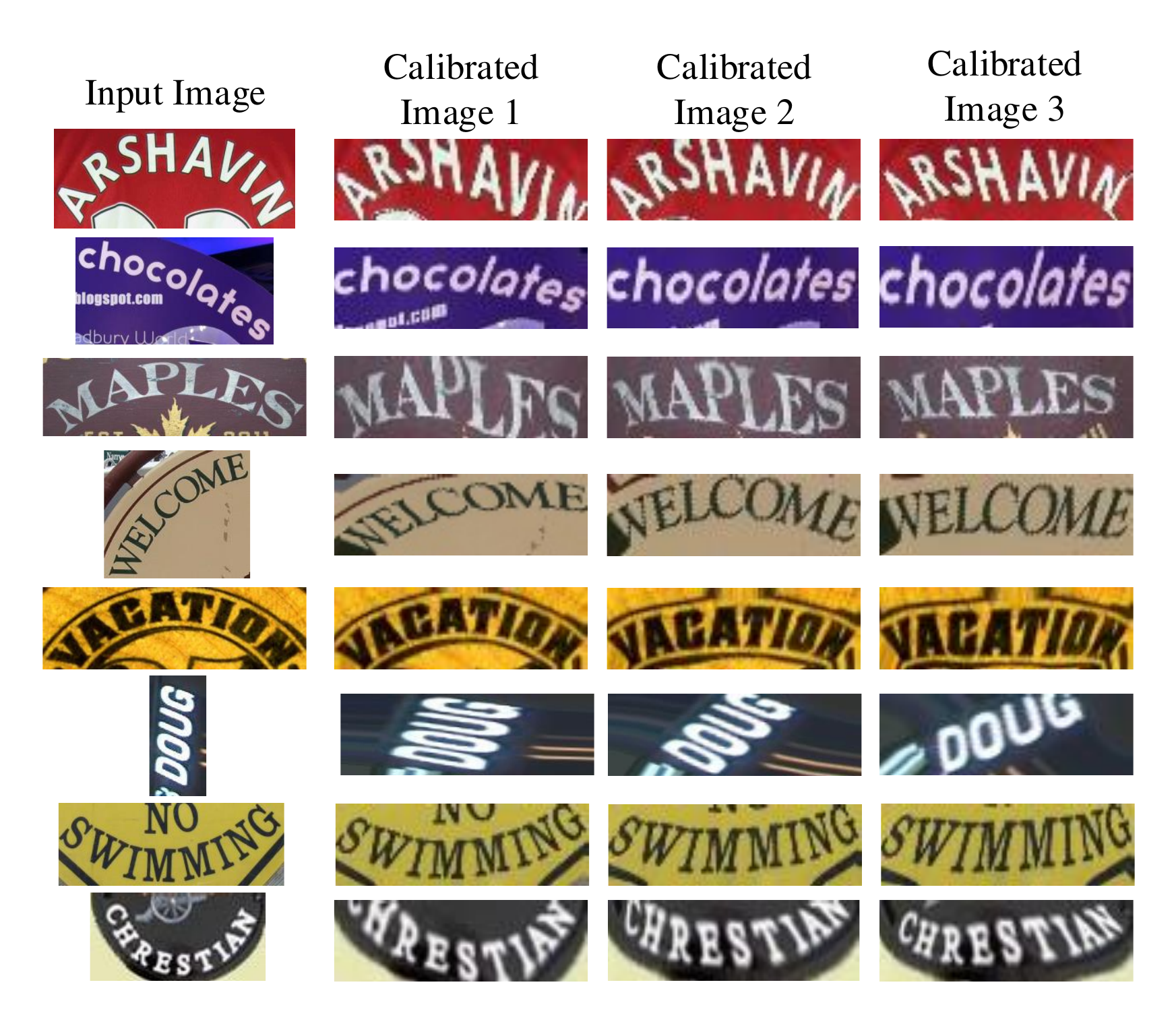}
\caption{Visualizations of the calibrated images during iterative refinements.}
\label{fig:picture003}
\end{figure}

Recognizing irregular text is very challenging, due to the various character placements. To validate the effectiveness of our method, we evaluate RCN on several irregular benchmarks and summarize the results in Table \ref{tab:001}. The network with a single calibration is taken as the baseline method. As observed, our method achieves significant improvements than the baseline, and consistently outperforms the other approaches by a large margin. It is worth noting that we do not reproduce the reported results in \cite{shi2018aster}, by comparing the baseline and \cite{shi2018aster}. Nevertheless, we still achieve better performance on all benchmarks, which suggests the significance of our method. Especially, we outperform \cite{shi2018aster} by a margin of 9 percentages on CUTE80. Besides, we find that the performance gains on the curved text benchmarks are more significant than the perspective text benchmarks. The distortions of curved text are more serious and hard to model, so existing methods perform worse on curved text. By contrast, our approach can effectively calibrate severely distorted text, and hence obtains much promotion on curved text benchmarks. It also should be pointed out that the RCN not only achieves much better calibrations, but also has no any extra parameter. As shown in Figure \ref{fig:picture003}, our RCN is capable of calibrating the irregular text with various degrees of deformation, including the nearly vertical text. Compared with \cite{cheng2017arbitrarily}, we do not destroy the aspect ratio of text, and thus the characters have no deformation. We also report the recognition performance on Total-Text that has not been recorded in previous literature. Our method achieves promising results on both multi-oriented and curved text collections. Furthermore, with the calibrated text from the last iteration, any kind of recognition network can be exploited. If using the focusing attention mechanism in \cite{cheng2017focusing} and the edit probability in \cite{bai2018edit}, the performance can be further improved.

\subsection{Performance on Regular Benchmarks}

We also conduct experiments on regular benchmarks. Most samples in these datasets are regular text, but irregular text also exists. We report our results in Table \ref{tab:002}. Shi \emph{et al.} \cite{shi2018aster} corrected the results on SVT in their released code and we record the updated results. The baseline is the network with a single calibration as in \cite{shi2018aster}. Compared with the baseline method, the RCN significantly improves the recognition performance. We can see that the RCN performs best on IIIT5k in lexicon-free setting. It is observed that IIIT5k contains many curved text, which demonstrates the advantage of RCN in dealing with severely distorted text. In the lexicon-based scenario, we achieve the best results on SVT and IIIT5k, but slightly fall behind \cite{cheng2017focusing,jaderberg2016reading} on IC03 and \cite{bai2018edit} on IC13. However, \cite{bai2018edit} applied the specially designed edit probability to train their networks, while we only use the traditional frame-wise loss. Besides, \cite{jaderberg2016reading} benefited from a pre-defined 90k lexicon and only recognized the words in its dictionary. It also should be remarked that \cite{cheng2017focusing} and \cite{yang2017learning} used the extra character bounding box annotations. By contrast, our method just requires the textual labels, which saves a lot of resources.

\section{Conclusion}

In this paper, we propose a Recurrent Calibration Network (RCN) for irregular text recognition. We divide the calibration process into multiple progressive steps to relieve the calibration difficulty of each step. 
Besides, the recurrent structure makes our network has the same parametric capacity as the network with a single spatial transformation. Moreover, we design a fiducial-point refinement structure to track and transmit the coordinates of fiducial points, instead of propagating the calibrated images. Therefore, our network is able to effectively keep the integrity of text during iterative calibrations. The lost information caused by inaccurate transformation can be recovered in subsequent processes. Furthermore, the calibration network and recognition network are jointly trained under the same objective for text recognition. Thus, the text is gradually calibrated in the direction that is more beneficial to recognition. Extensive experiments conducted on challenging benchmarks verify the effectiveness of our method, especially on irregular datasets.

{\small
\bibliographystyle{ieee}
\bibliography{egbib}
}

\end{document}